\documentclass{article}

\PassOptionsToPackage{sort&compress}{natbib}
\usepackage[final]{corl_2026}
\makeatletter
\renewcommand{\@conferencelocation}{Austin, Texas, USA}
\makeatother
\usepackage[utf8]{inputenc}
\usepackage{amsmath,amssymb,mathtools}
\usepackage{graphicx}
\usepackage{capt-of}
\usepackage{booktabs,longtable,array}
\usepackage{makecell}
\usepackage{wrapfig}
\usepackage{placeins}
\usepackage{url}
\hypersetup{
  pdftitle={PACE: Temporal-Role Tactile Conditioning for Chunk-Based Insertion Policies},
  pdfauthor={Zhiyuan Chen, Shixiong Xu, Shuai Chen, Yangwei You}
}
\graphicspath{{}}
\newcommand{\linkto}[2]{\hyperref[#1]{#2}}
\newcommand{\figref}[1]{\hyperref[#1]{Figure~\ref*{#1}}}
\newcommand{\tabref}[1]{\hyperref[#1]{Table~\ref*{#1}}}
\newcommand{\secref}[1]{\hyperref[#1]{Section~\ref*{#1}}}
\newcommand{\appref}[1]{\hyperref[#1]{Appendix~\ref*{#1}}}
\newcommand{\eqnref}[1]{\hyperref[#1]{Equation~(\ref*{#1})}}
\makeatletter
\newcommand{\wrappedtablecaption}[1]{%
  \setlength{\abovecaptionskip}{\@nipsbelowcaptionskip}%
  \setlength{\belowcaptionskip}{\@nipsabovecaptionskip}%
  \caption{#1}%
}
\makeatother

\title{PACE: Temporal-Role Tactile Conditioning for Chunk-Based Insertion Policies}
\author{
  Zhiyuan Chen\thanks{Corresponding author.} \quad
  Shixiong Xu \quad Shuai Chen \quad Yangwei You\\
  Beijing Xiaomi Robot Technology Co., Ltd.
}
\begin{document}
\maketitle

\begin{minipage}{\linewidth}
\centering
\includegraphics[width=\linewidth,totalheight=0.23\textheight,keepaspectratio]{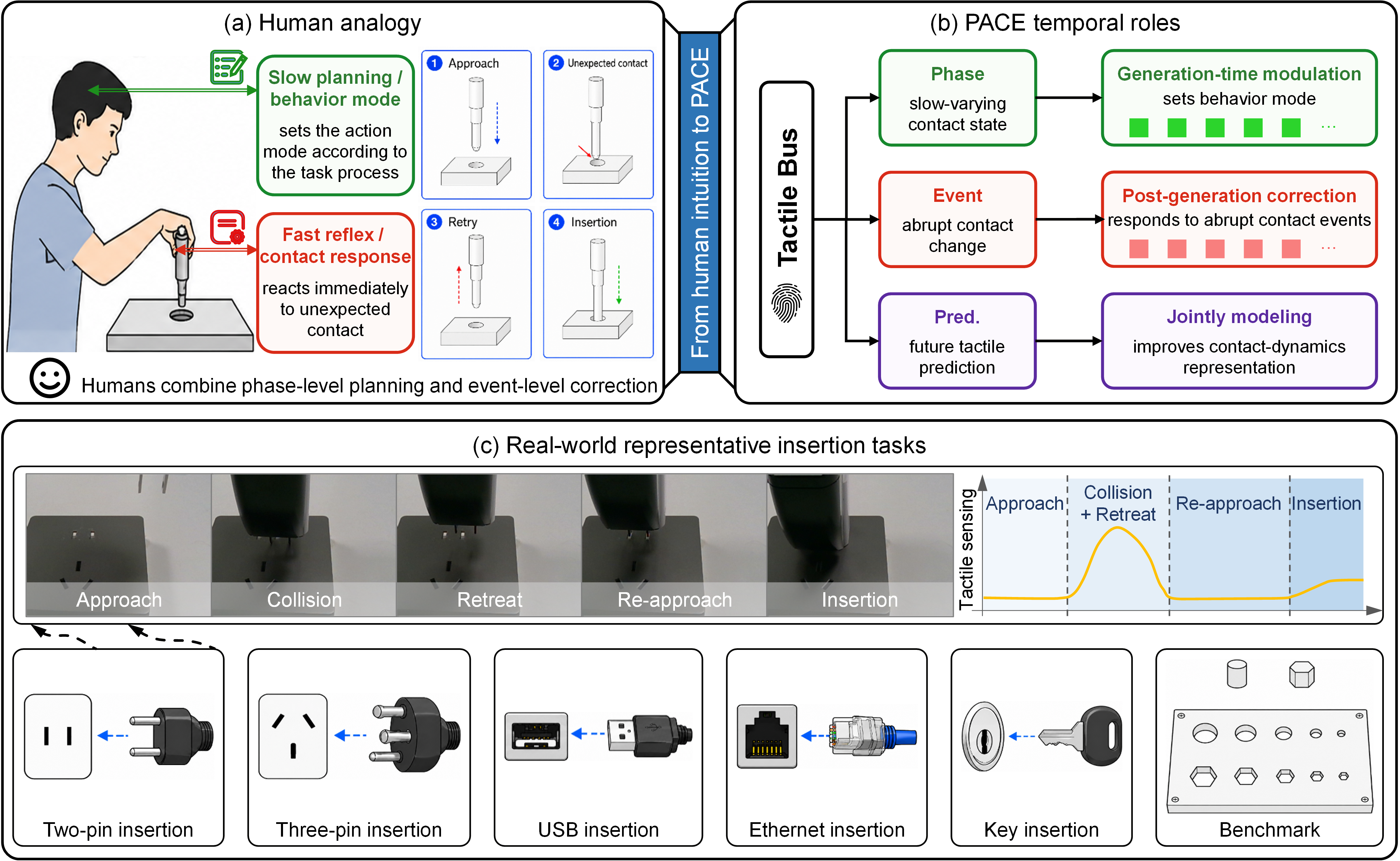}
\captionof{figure}{\textbf{PACE.} A temporal-role tactile conditioning framework that organizes one tactile stream into three roles: contact state conditions generation via affine modulation, contact events correct the chunk via gated residual, and contact-consequence prediction shapes representations during training.}
\label{fig:teaser}
\end{minipage}

\begin{abstract}
Contact-rich insertion requires robots to combine visual alignment with tactile adjustment. However, existing visuotactile chunk-based policies collapse tactile feedback into a single generation-time fusion pathway, leaving persistent contact states, abrupt contact events, and future contact consequences implicitly modeled in a single representation. We present \textbf{PACE} (Phase-Aware Contact-Event Conditioning), a temporal-role tactile conditioning framework for chunk-based insertion policies. PACE organizes a shared tactile stream into three temporal roles: persistent contact states condition action-chunk generation, abrupt contact events correct the generated chunk, and contact-consequence prediction shapes action representations during training. This design preserves chunk-level action continuity while assigning distinct pathways to different temporal roles of contact feedback. Across five real-robot insertion tasks, PACE achieves an average success rate of 90.7\%, outperforming vision-only and generation-time tactile-fusion baselines by over 44 percentage points; on the ManipulationNet benchmark, PACE maintains 88.7\% success across six geometry-clearance configurations, including an industrial-grade 0.1-mm clearance. Ablations confirm the complementary contributions of the three temporal roles, mechanistic analyses show that PACE separates normal insertion progress from contact events requiring immediate correction, and cross-signal and cross-architecture experiments indicate that this temporal-role principle extends across different physical feedback signals and policy backbones. Videos and materials are available at \url{https://pace-insertion.github.io/}.
\end{abstract}


\section{Introduction}\label{sec:introduction}

Precision insertion requires robots to combine global alignment with local contact adjustment. Vision can provide object pose, target location, and coarse motion direction in free space. After contact occurs, however, the end effector and the workpiece often occlude the most informative regions. Small misalignments, rim collisions, and local contact changes can be difficult to infer reliably from images in time, while they already produce clear feedback in tactile or force signals. Physical feedback is therefore critical for post-contact local adjustment in small-clearance connector assembly, peg-in-hole insertion, key insertion, and related tasks \citep{ref36,ref37,ref39,ref41,ref53}.

Classical methods for contact-rich manipulation often rely on impedance and compliance control \citep{ref1,ref3,ref4}, hybrid force/position control \citep{ref2}, or operational-space force control \citep{ref5}. These methods can be highly effective when geometry and control parameters are carefully specified, but they generalize poorly to diverse real-world insertion scenarios with varying geometries. End-to-end imitation learning provides a more flexible alternative. Among such methods, action-chunking policies \citep{ref14,ref15,ref20,ref22,ref24} have become a common design: a policy predicts a short sequence of future actions in a single forward pass, improving motion smoothness, amortizing inference cost, and maintaining action consistency.

After contact occurs, insertion progress is largely determined by how physical interaction evolves over time. In human insertion, contact is not uniformly interpreted: slowly varying contact state signals the current manipulation phase and sets behavioral modes \citep{ref45,ref47}, abrupt contact transients trigger immediate corrective reflexes \citep{ref46}, and anticipated contact consequences support efficient motor planning \citep{ref45}. Tactile feedback in contact-rich insertion is therefore not merely an additional sensing modality, but a control signal with distinct temporal roles.

Existing visuotactile policies typically fuse tactile feedback as a single generation-time input to the action generator. This single-path design forces contact state, contact events, and contact consequences to be learned implicitly within one representation. The question is therefore not merely whether to use tactile feedback, but how to organize it according to its temporal roles.

We propose \textbf{PACE} (Phase-Aware Contact-Event Conditioning), a temporal-role tactile conditioning framework built on an Action Chunking with Transformers (ACT)-style chunk policy \citep{ref14}. PACE organizes the same tactile stream into three learnable temporal roles: \textbf{contact-state conditioning} guides action-chunk generation, \textbf{contact-event correction} corrects the generated chunk, and \textbf{contact-consequence prediction} shapes action representations during training.

We evaluate PACE on the ManipulationNet benchmark and five real insertion tasks, covering a standard peg-in-hole benchmark as well as real connector insertion with keys, plugs, USB, and Ethernet objects. Our main contributions are:

\begin{itemize}
\item \textbf{The PACE tactile conditioning framework.} We propose a tactile conditioning framework that assigns the same tactile stream to three learnable temporal roles within chunk-based control: contact-state conditioning for generation, contact-event correction of the generated chunk, and contact-consequence prediction during training.
\item \textbf{Systematic real-robot validation, ablations, and mechanistic analysis.} Across five real insertion tasks, PACE achieves an average success rate of 90.7\%, outperforming vision-only policies and generation-time tactile-fusion baselines by over 44 percentage points (pp); on ManipulationNet, PACE maintains 88.7\% success, including 0.1 mm clearance. Ablations and mechanistic analyses show that the three temporal roles are complementary and produce distinct phase/event signatures.
\item \textbf{Cross-signal and cross-architecture generality.} PACE transfers to Diffusion Policy \citep{ref15} and operates with end-effector force/torque or joint currents instead of the default tactile sensor, indicating that temporal-role organization is not tied to the tested tactile sensor or ACT backbone.
\item \textbf{Controlled validation of force-feedback demonstrations.} Using the same device, the same operator, and the same task setup while only toggling force-feedback rendering, we show that force-feedback teleoperation produces more consistent demonstrations with natural contact-responsive behaviors, providing higher-quality data for learning PACE's three pathways.
\end{itemize}

\section{Related Work}\label{sec:related_work}

\subsection{Precision Insertion and Contact-Rich Manipulation}\label{sec:related_insertion}

Precision insertion typically includes approach, first contact, search, alignment, and stable insertion, along with rim collisions, local jamming, slip, and abrupt force changes. Visual alignment can therefore be insufficient: sub-millimeter pose errors may be hidden by occlusion, specular reflection, or local geometry, while tactile and force feedback already expose the contact state.

Classical methods use impedance and compliance control \citep{ref1,ref3,ref4}, hybrid force/position control \citep{ref2}, and operational-space force control \citep{ref5}, but depend strongly on geometric priors and parameter tuning. Learning-based methods acquire contact behavior from data, including guided policy search and reinforcement-learning assembly \citep{ref6,ref7,ref10}, tight insertion and industrial connector learning \citep{ref12,ref13}, visuotactile behavior cloning \citep{ref41,ref44,ref54}, and force/tactile-aware policies \citep{ref11,ref43,ref52,ref53}. These works motivate contact sensing, but do not organize contact signals according to their temporal roles.

\subsection{Action-Chunk Imitation Learning}\label{sec:related_chunking}

Action-chunking policies predict multiple future actions per query and underpin ACT \citep{ref14}, Diffusion Policy \citep{ref15}, robotic foundation models such as \(\pi_0\) \citep{ref24} and OpenVLA \citep{ref32}, and recent action-tokenization/VLA systems \citep{ref16,ref18,ref19}. Chunks improve smoothness and short-horizon consistency, but create a temporal gap between the latest observation and already generated actions, which is critical when contact changes inside a chunk.

Recent work improves closed-loop chunk execution by addressing inter-chunk continuity or refresh timing: RTC \citep{ref20} and REMAC \citep{ref22} target real-time execution and smoothness, DCDP \citep{ref21} refreshes chunks after visual changes, and RDP combines slow latent diffusion with fast tactile-conditioned action decoding during execution \citep{ref23}. PACE builds on this family of chunk-based policies and focuses on how tactile feedback should be organized within the policy pipeline according to its temporal roles.

\subsection{Visuotactile Policy Learning}\label{sec:related_visuotactile}

Visuotactile fusion methods range from token concatenation \citep{ref38,ref41} to cross-modal attention \citep{ref35,ref40}, dynamic modality weighting \citep{ref43,ref44}, humanoid touch-conditioned manipulation \citep{ref42}, and force-aware action generation \citep{ref53,ref56}. Predictive tactile objectives and visuotactile world models further improve contact representation and planning \citep{ref27,ref55,ref57,ref58}, while contrastive pretraining inspired by vision-language alignment \citep{ref25} has been adopted to learn general-purpose tactile features \citep{ref60}. These methods mainly treat tactile feedback as a modality fused during generation.

How tactile feedback should be organized according to its temporal roles remains underexplored. Existing designs primarily fuse tactile feedback as a single generation-time input, without explicitly distinguishing contact state, contact events, and contact consequences. PACE addresses this gap by decomposing the same contact stream into three explicit temporal-role pathways.

\section{Method: PACE}\label{sec:method}

\subsection{Problem Definition and Method Overview}\label{sec:problem_overview}

We study chunk-based imitation learning for contact-rich insertion. The policy receives global RGB, wrist RGB, proprioception, and low-dimensional tactile feedback, and an ACT-style Conditional Variational Autoencoder (CVAE)-Transformer backbone \citep{ref14,ref28,ref29} predicts a multi-step action chunk at each query. The default tactile input is a 6D tactile-force signal; \secref{sec:cross_signal} tests end-effector force/torque and joint currents.

The central challenge is that single-path tactile fusion forces a policy to learn contact state, contact events, and contact consequences implicitly within one representation. Rather than introducing a new tactile encoder, PACE changes where and when tactile information acts in a chunk-based policy. It provides three explicit pathways, each specialized for one temporal role of contact feedback.

\begin{figure}[!htbp]
\centering
\includegraphics[width=0.90\linewidth,keepaspectratio]{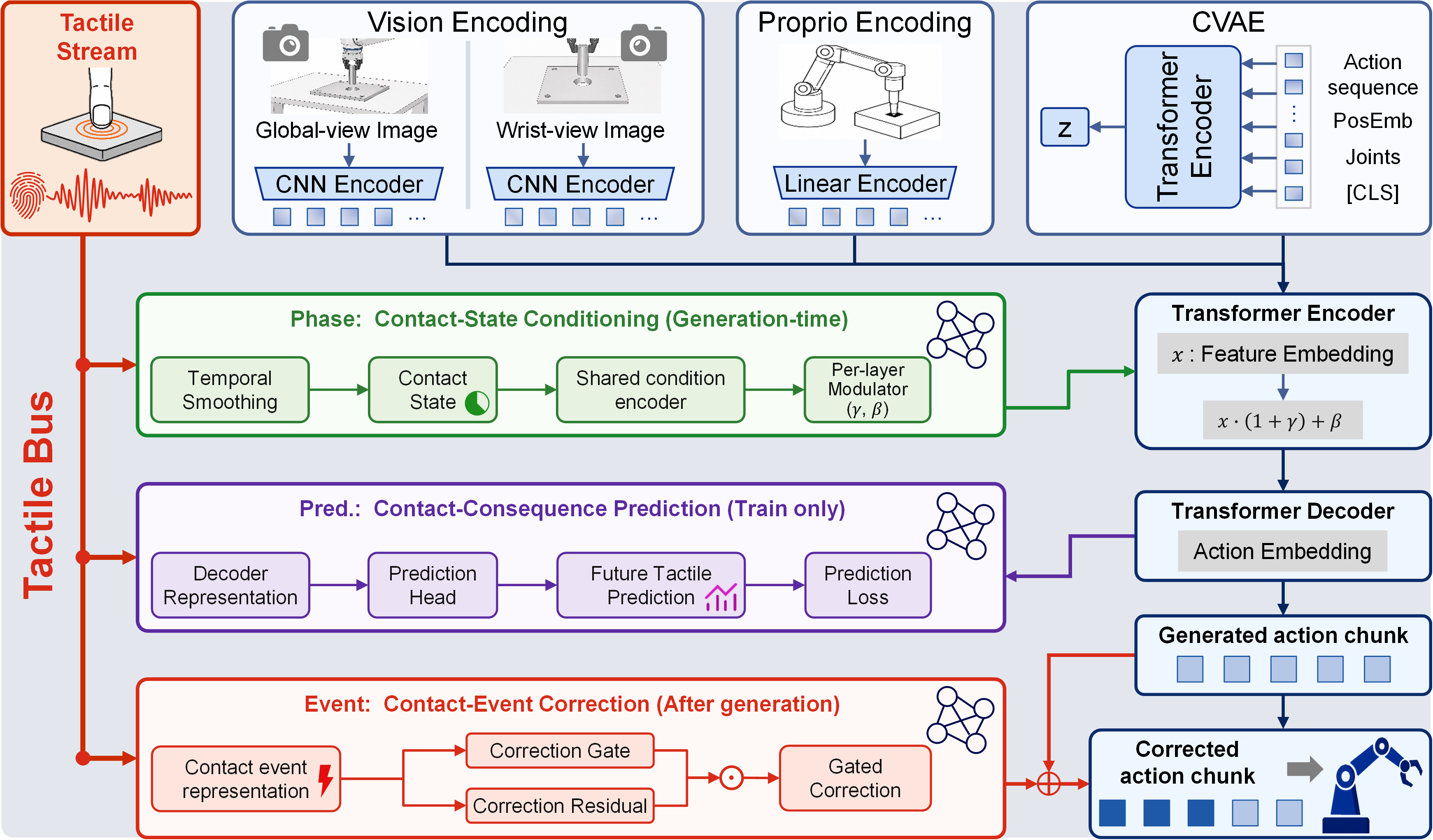}
\caption{\textbf{Overview of PACE.} Multimodal observations enter specialized encoders; a shared tactile bus feeds three temporal-role pathways. Phase: smoothed contact state modulates Transformer encoder features via layer-wise affine conditioning (generation-time). Pred.: decoder representations are supervised to predict future contact changes (training only). Event: contact-event cues produce a gated residual that corrects the generated action chunk (post-generation). The backbone is an ACT-style CVAE-Transformer.}
\label{fig:architecture}
\end{figure}

PACE addresses this challenge by assigning the same contact stream to three temporal roles (\figref{fig:architecture}). \textbf{Phase} uses slowly varying contact state to condition action generation, \textbf{Event} uses abrupt contact changes to correct the generated chunk, and \textbf{Pred.} predicts future contact changes during training. The following sections describe the three roles; \appref{app:network} gives architectural details.

\subsection{Contact-State Conditioning for Action-Chunk Generation}\label{sec:phase}

Fine insertion proceeds through phases such as approach, first contact, search, and stable insertion, each requiring different action tendencies. Phase captures this slowly varying state with an exponential moving average:
\begin{equation}
\mathbf{w}^{\text{state}}_t = (1-\alpha)\mathbf{w}^{\text{state}}_{t-1} + \alpha \mathbf{w}_t.
\label{eq:phase_ema}
\end{equation}
Here, \(\mathbf{w}_t\) is the raw tactile reading and \(\alpha\) is the smoothing rate. At query time \(s\), the smoothed state suppresses transient noise and preserves phase-related contact information before the policy commits to a new action chunk. PACE conditions Transformer encoder features through feature-wise affine modulation \citep{ref33}:
\begin{equation}
(\gamma^l_s, \beta^l_s) = \phi^l_{\text{phase}}(\mathbf{w}^{\text{state}}_s), \quad \widetilde{\mathbf{h}}_l = (1+\gamma^l_s) \odot \mathrm{LN}(\mathbf{h}_l) + \beta^l_s.
\label{eq:phase_affine}
\end{equation}
Here, \(\phi^l_{\text{phase}}\) is a learned contact-state modulation map for encoder layer \(l\); \(\mathrm{LN}\) denotes layer normalization \citep{ref34}; \(\mathbf{h}_l\) is the encoder feature, and \(\gamma^l_s,\beta^l_s\) are the resulting affine modulation parameters. This lets contact state rescale and shift encoder features so that generation can switch behavior modes, for example from cautious probing to steady insertion.

\subsection{Contact-Event Correction of the Generated Chunk}\label{sec:event}

Phase handles slow contact state, but collision, slip, and jamming appear as short-time contact changes. Event therefore constructs a short-window contact-change cue and applies a gated residual correction to the front of the generated action chunk. This gives the policy an explicit pathway for contact-change-driven local adaptation, rather than forcing the action generator to encode all contact transients implicitly.
\begin{equation}
\bar{\mathbf{w}}_s = \mathcal{N}(\mathbf{w}_s), \quad
\Delta \mathbf{w}_s = \bar{\mathbf{w}}_s - \bar{\mathbf{w}}_{s-T_w+1}, \quad
\mathbf{e}_s = [\Delta \mathbf{w}_s, \|\Delta \mathbf{w}_s\|_2],
\label{eq:event_cue}
\end{equation}
where \(\mathcal{N}(\cdot)\) applies training-set z-score normalization, \(T_w\) is the tactile history window, and \(\mathbf{e}_s\) concatenates the contact-change vector with its magnitude.
\begin{equation}
\mathbf{g}_s = \sigma(\phi_{\text{gate}}(\mathbf{e}_s)), \quad
\mathbf{R}_{s,0:h_{\text{corr}}-1} =
\phi_{\text{res}}(\mathbf{W}_s) \in \mathbb{R}^{h_{\text{corr}}\times D_a},
\label{eq:event_gate_residual}
\end{equation}
where the gate \(\mathbf{g}_s\) is produced by the learned event gate \(\phi_{\text{gate}}\), the learned residual map \(\phi_{\text{res}}\) predicts \(\mathbf{R}_{s,0:h_{\text{corr}}-1}\) from the tactile history \(\mathbf{W}_s\), and \(D_a\) is the action dimension.
\begin{equation}
\mathbf{A}^{\text{corr}}_{s,0:h_{\text{corr}}-1}
= \mathbf{A}^{\text{gen}}_{s,0:h_{\text{corr}}-1}
+ \mathbf{g}_s \odot \mathbf{R}_{s,0:h_{\text{corr}}-1}.
\label{eq:event_corrected_chunk}
\end{equation}
Here, \(\mathbf{A}^{\text{gen}}_s\) is the generated action chunk, \(\mathbf{A}^{\text{corr}}_s\) is the corrected chunk, and \(h_{\text{corr}}=5\) is the corrected prefix length (0.25 s under a 20 Hz control rate). The gate is broadcast over the corrected prefix, so only the near-term actions are locally adjusted while the longer-horizon chunk structure is preserved. At each policy query \(s\), Event uses the latest tactile window to correct the action chunk generated by that same forward pass; the corrected chunk is then passed to the execution mechanism.

\subsection{Contact-Consequence Prediction}\label{sec:pred}

Behavior cloning may fit visual-action correlations without representing how actions change contact. Pred. therefore supervises decoder hidden states \(\mathbf{z}_{s:s+H-1}\) to predict future normalized tactile increments:
\begin{equation}
\mathcal{L}_{\text{pred}} =
\left\|\phi_{\text{pred}}(\mathbf{z}_{s:s+H-1})
- \Delta\mathbf{W}^{\text{gt}}_{s:s+H-1}\right\|_2^2,
\label{eq:pred_loss}
\end{equation}
where the decoder states \(\mathbf{z}_{s:s+H-1}\) are supervised by a learned prediction head \(\phi_{\text{pred}}\); the target \(\Delta\mathbf{W}^{\text{gt}}_{s:s+H-1}\) is the sequence of normalized tactile increments, with \(\Delta\mathbf{w}^{\text{gt}}_t=\mathcal{N}(\mathbf{w}_t)-\mathcal{N}(\mathbf{w}_{t-1})\). Each future action query is therefore supervised not only to match the expert action, but also to encode contact changes around the corresponding control step. The prediction head is removed during inference.

\subsection{Training Objective}\label{sec:objective}

Phase, Event, and Pred. are trained end to end and act on different stages of the pipeline: encoder features, generated chunk actions, and decoder hidden states. The total objective is
\begin{equation}
\mathcal{L} =
\mathcal{L}_{\text{act}}
+ \lambda_{\text{kl}}\mathcal{L}_{\text{KL}}
+ \lambda_{\text{pred}}\mathcal{L}_{\text{pred}}.
\label{eq:total_loss}
\end{equation}
The action term \(\mathcal{L}_{\text{act}}\) is the L1 imitation loss on the chunk after Event correction: the first \(h_{\text{corr}}\) steps use \(\mathbf{A}^{\text{corr}}_{s,0:h_{\text{corr}}-1}\), and the remaining steps use \(\mathbf{A}^{\text{gen}}_{s,h_{\text{corr}}:H-1}\). \(\mathcal{L}_{\text{KL}}\) is the ACT/CVAE latent regularizer, and \(\lambda_{\text{kl}}=10\), \(\lambda_{\text{pred}}=0.05\) weight the auxiliary terms. \linkto{app:network}{Appendices~A} and~\linkto{app:training}{C} provide network and training details.

During inference, only Phase and Event remain. Compared with the 83.9 M-parameter ACT backbone, PACE adds 0.841 M parameters during training, of which 0.575 M remain at inference, with less than 0.3 ms additional latency.

\section{Experiments and Analysis}\label{sec:experiments}

The experiments test four claims: PACE improves insertion success over vision-only and generation-time tactile-fusion baselines; Phase, Event, and Pred. each contribute to performance; the learned pathways exhibit distinct phase/event signatures; and the temporal-role design generalizes across signal sources and policy backbones.

\subsection{Experimental Setup}\label{sec:setup}

\begin{figure}[!htbp]
\centering
\includegraphics[width=0.72\linewidth]{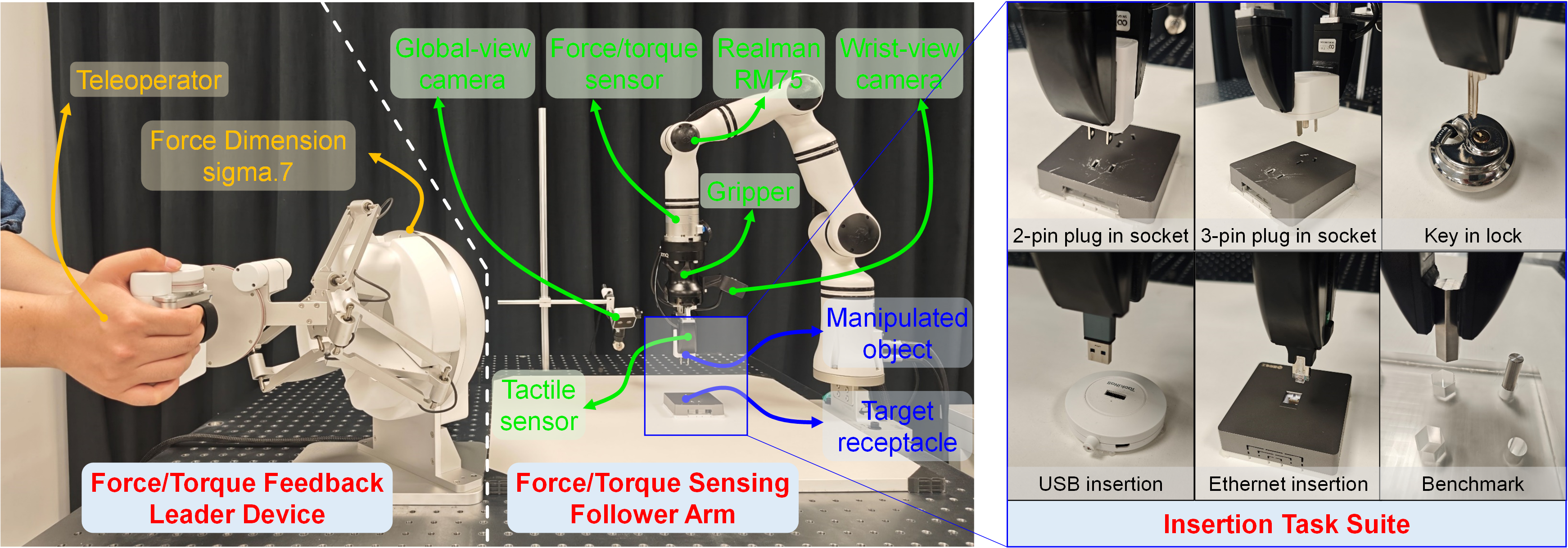}
\caption{\textbf{Experimental setup.} Left: force-feedback teleoperation system (Force Dimension sigma.7 leader, robot follower). Right: five real insertion tasks and the ManipulationNet benchmark.}
\label{fig:hardware}
\end{figure}

We evaluate PACE on the hardware platform and task objects shown in \figref{fig:hardware}: a Realman RM75 arm with a Robotiq Hand-E gripper, two RealSense D405 cameras, a Vitai GF225 fingertip tactile sensor, end-effector force/torque sensing, and joint-current logging. Force-feedback demonstrations are collected with a Force Dimension sigma.7 that renders calibrated contact wrench to the operator \citep{ref48,ref49,ref50,ref51}, yielding collision retreat, search adjustment, and stable insertion behaviors. Each task has about 160 demonstrations of roughly 300 steps; \appref{app:collection} details the collection system.

We use two evaluation suites. ManipulationNet \citep{ref39} provides the NIST ATB peg-in-hole benchmark with cylindrical/hexagonal pegs at 3, 1, and 0.1 mm clearances. The real-robot suite contains Key, 2-pin plug, 3-pin plug, USB, and Ethernet insertion, covering orientation constraints, hidden pins, multi-point contact, small-clearance sensitivity, and sustained-force insertion. Each method is evaluated with 30 independent rollouts per real task under 25 mm \(\times\) 25 mm target randomization; success requires stable insertion or functional connection.

The five real tasks stress different contact ambiguities. Key and USB require precise orientation and are sensitive to rim collisions, testing event-driven correction after abrupt contact. 2-pin hides critical geometry from the wrist view during approach, forcing post-contact search based on physical feedback, while 3-pin adds a three-point geometric constraint. Ethernet contains an elastic latch and requires sustained insertion force, so increased force can indicate either normal progress or harmful jamming. These differences let us test whether PACE handles both slow contact-phase changes and abrupt local events.

\subsection{Baselines}\label{sec:baselines}

Baselines include vision-only ACT \citep{ref14}, vision-only Diffusion Policy \citep{ref15}, fine-tuned \(\pi_0\) \citep{ref24,ref31}, and two visuotactile fusion variants: tactile cross-attention and tactile concatenation. All methods use the same force-feedback demonstrations, visual inputs, action space, and evaluation protocol; they differ primarily in policy architecture and contact-feedback usage. Thus, the comparisons in \linkto{tab:real_results}{Tables~1}--\linkto{tab:ablations}{2} are not confounded by demonstration source; the force-feedback study in \secref{sec:ff_quality} changes the demonstration source while keeping the PACE architecture fixed.

An additional targeted comparison evaluates RDP \citep{ref23}, a reactive visuotactile baseline, on 2-pin and Key using the same demonstrations and evaluation protocol; details appear in \appref{app:rdp}.

\subsection{Main Results}\label{sec:main_results}

This experiment examines whether PACE's temporal-role decomposition provides consistent advantages over vision-only policies and generation-time tactile-fusion baselines across diverse insertion geometries. The real-robot comparison uses the same demonstrations, visual inputs, action space, and 30 evaluation rollouts per task for all methods. ManipulationNet further evaluates PACE over six geometry-clearance configurations, with 25 rollouts per configuration.

\begin{table}[t]
\centering
\renewcommand{\thetable}{1(a)}
\caption{Quantitative comparison on real insertion tasks.}
\label{tab:real_results}
\begin{tabular*}{\linewidth}{@{\extracolsep{\fill}}lcccccc@{}}
\toprule
Method & Key & 2-pin & 3-pin & USB & Ethernet & Avg. \\
\midrule
Vision-only ACT & 4/30 & 4/30 & 2/30 & 1/30 & 4/30 & 10.0\% \\
Vision-only DP & 5/30 & 6/30 & 4/30 & 3/30 & 5/30 & 15.3\% \\
\(\pi_0\) & 6/30 & 8/30 & 4/30 & 4/30 & 6/30 & 18.7\% \\
ACT + Tactile Cross-Attn & 10/30 & 15/30 & 11/30 & 9/30 & 14/30 & 39.3\% \\
ACT + Tactile Concat & 13/30 & 17/30 & 12/30 & 13/30 & 14/30 & 46.0\% \\
\textbf{PACE} & \textbf{27/30} & \textbf{29/30} & \textbf{28/30} & \textbf{27/30} & \textbf{25/30} & \textbf{90.7\%} \\
\bottomrule
\end{tabular*}
\end{table}

\begin{wraptable}{r}{0.40\linewidth}
\centering
\renewcommand{\thetable}{1(b)}
\renewcommand{\theHtable}{1b}
\wrappedtablecaption{ManipulationNet peg-in-hole results.}
\label{tab:manip_pace}
\begin{tabular}{@{}lccc@{}}
\toprule
Peg & 3 mm & 1 mm & 0.1 mm \\
\midrule
Cyl. & 25/25 & 22/25 & 21/25 \\
Hex. & 24/25 & 22/25 & 19/25 \\
\bottomrule
\end{tabular}
\end{wraptable}
\setcounter{table}{1}

\noindent
As shown in \linkto{tab:real_results}{Table~1(a)}, PACE reaches 90.7\% average success, more than 44 pp above the strongest baseline. Vision-only policies (ACT 10.0\%, DP 15.3\%, \(\pi_0\) 18.7\%) struggled to resolve post-contact ambiguity despite increasingly capable backbones. Generation-time tactile fusion (Concat 46.0\%, Cross-Attn 39.3\%) improves over vision-only policies but still treats tactile feedback as a single generation-time input; notably, cross-attention does not outperform simple concatenation, suggesting that more complex generation-time fusion mechanisms do not necessarily improve performance when tactile feedback is treated as a single-path input. PACE (90.7\%) addresses this by assigning distinct temporal roles to Phase, Event, and Pred. We additionally compare PACE with RDP on 2-pin and Key using the same demonstrations and evaluation protocol. PACE achieves 29/30 and 27/30 successes, versus RDP's 19/30 and 12/30, respectively (\appref{app:rdp}). \linkto{tab:manip_pace}{Table~1(b)} complements the real-task results with standardized peg-in-hole evaluation. As shown in \linkto{tab:manip_pace}{Table~1(b)}, PACE also remains robust on ManipulationNet at the tightest 0.1 mm clearance. ManipulationNet baselines are reported in \appref{app:manip_baselines}.

Wilson 95\% confidence intervals for the 30-rollout real-robot comparisons are reported in \appref{app:statistical_reliability}.

\subsection{Structural and Design Ablations}\label{sec:ablations}

\begin{table}[!htbp]
\centering
\caption{Ablation study on PACE design choices.}
\label{tab:ablations}
\begin{tabular}{@{}lcccccc@{}}
\toprule
Method & Key & 2-pin & 3-pin & USB & Eth. & Avg. \\
\midrule
PACE w/o Pred. & 25/30 & 27/30 & 26/30 & 26/30 & 24/30 & 85.3\% \\
PACE w/o Phase & 24/30 & 26/30 & 21/30 & 23/30 & 24/30 & 78.7\% \\
PACE w/o Event & 20/30 & 22/30 & 18/30 & 19/30 & 21/30 & 66.7\% \\
Phase-only & 17/30 & 20/30 & 16/30 & 18/30 & 19/30 & 60.0\% \\
\textbf{PACE} & \textbf{27/30} & \textbf{29/30} & \textbf{28/30} & \textbf{27/30} & \textbf{25/30} & \textbf{90.7\%} \\
\bottomrule
\end{tabular}
\end{table}

This ablation examines whether the three temporal roles are complementary. All variants use the same training data, network scale, and evaluation protocol. Phase-only keeps generation-time tactile affine modulation but removes Event and Pred.

As shown in \linkto{tab:ablations}{Table~2}, removing Event causes the largest drop (90.7\% → 66.7\%), with the steepest degradation on 3-pin ($-$10) and USB ($-$8) --- tasks where frequent rim collisions demand immediate correction of the generated chunk. Removing Phase causes a moderate drop (→ 78.7\%), mainly hurting 3-pin ($-$7) and USB ($-$4) where multi-stage search relies on phase-dependent behavior modulation. Removing Pred. yields a smaller but consistent drop (→ 85.3\%), indicating that predictive training supervision improves representation quality without being critical at inference time. Phase-only reaches 60.0\%, indicating that contact-state conditioning alone is insufficient.

These ablations show that Phase, Event, and Pred. make complementary contributions. A further short-chunk control (Phase-only with \(H=10\), reported in \appref{app:chunk_length}) drops to 49.3\%, suggesting that simply increasing query frequency is insufficient to recover the benefits of event correction: shorter chunks weaken temporal coherence and sustained-force behavior, while Event keeps the long chunk structure and adds targeted local correction.

\subsection{Mechanistic Analysis}\label{sec:mechanistic}

This analysis examines whether Phase and Event produce distinct temporal signatures aligned with contact phases and contact events. We use a representative rollout for temporal inspection and aggregate rollout statistics for quantitative support.

\begin{figure}[!htbp]
\centering
\includegraphics[width=\linewidth,totalheight=\textheight,keepaspectratio]{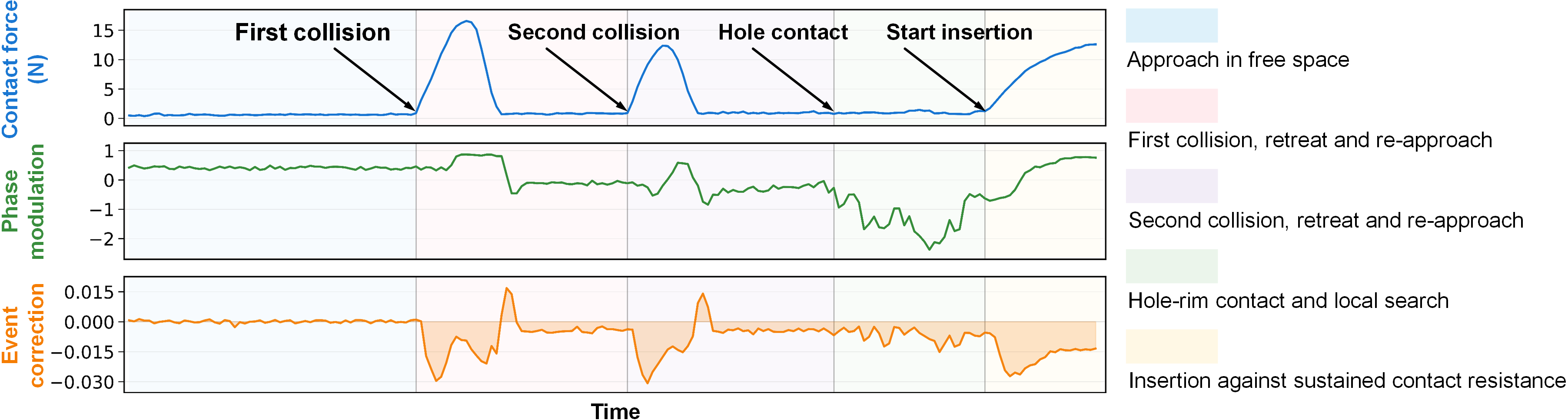}
\caption{Mechanistic visualization of PACE's temporal-role separation during insertion. Top: calibrated end-effector contact force (N). Middle: Phase pathway output, visualized as the second principal component of the feature-wise affine modulation parameters ($\gamma, \beta$) (a.u.). Bottom: Event pathway output, shown as the gate-weighted residual correction level relative to its pre-contact resting level (a.u.). Phase exhibits sustained shifts across different contact phases, whereas Event shows sharp excursions around collision events and does not simply increase with the sustained high force of stable insertion.}
\label{fig:mechanistic_temporal}
\end{figure}

\textbf{Temporal case study.} As shown in \figref{fig:mechanistic_temporal}, Phase modulation follows the broader contact phase: it remains relatively stable during approach, shifts after collision events, fluctuates during repeated search-and-retreat, and stabilizes during insertion. The Event correction level deviates sharply from its pre-contact baseline around collision or jamming events, but does not simply track the sustained force increase during stable insertion. This distinction matters because stable insertion can have high force without requiring correction, while abrupt contact changes at similar force levels may require immediate local updates. These results indicate that PACE does not learn a trivial mapping from larger force to larger correction; it separates normal contact progress from abrupt contact events that require local chunk correction.

\textbf{Aggregate statistics.} To test whether the single-rollout pattern generalizes across evaluation rollouts, we aggregate each of the 30 2-pin rollouts by contact-intensity bin, then compare Phase modulation norm across bins and Event correction level with contact-change rate using rollout-level summaries. Phase modulation differs significantly across bins (ANOVA, \(F=8.36, p<10^{-4}\), \(\eta^2=0.236\)), while Event correction correlates positively with contact-change rate (Pearson \(r=0.356, p<0.001\)). These results support the intended division of labor: Phase tracks slow contact progress, while Event updates actions near collision or jamming events.

\subsection{Cross-Signal-Source Generality}\label{sec:cross_signal}

This experiment tests cross-signal generality by replacing the default GF225 tactile-force signal with end-effector six-axis force/torque or joint currents. We evaluate on 2-pin and Key using the same training data and 30 evaluation rollouts per task as in Table 1. This matters because available contact-sensitive feedback signals differ across robot platforms.

\begin{wraptable}{r}{0.56\linewidth}
\centering
\wrappedtablecaption{Alternative physical feedback signals.}
\label{tab:signals}
\begin{tabular}{@{}lcccc@{}}
\toprule
Signal & Dim. & 2-pin & Key & Avg. \\
\midrule
Tactile (default) & 6 & 29/30 & 27/30 & 93.3\% \\
F/T & 6 & 27/30 & 27/30 & 90.0\% \\
Joint current & 7 & 28/30 & 25/30 & 88.3\% \\
\bottomrule
\end{tabular}
\end{wraptable}

\noindent
As shown in \linkto{tab:signals}{Table~3}, F/T and joint-current inputs remain close to the tactile default (93.3\%), reaching 90.0\% and 88.3\%. These results indicate that PACE is not limited to the default tactile sensor and can operate with alternative contact-sensitive signals.

\WFclear
\subsection{Cross-Architecture Validation}\label{sec:cross_arch}

This experiment tests whether PACE is specific to ACT by integrating the same Phase/Event/Pred. roles into Diffusion Policy \citep{ref15} and evaluating on 2-pin and Key. Phase conditions generation features, Event corrects the generated chunk, and Pred. supervises future contact changes.

\begin{wraptable}{r}{0.48\linewidth}
\centering
\wrappedtablecaption{Cross-architecture validation.}
\label{tab:dp_pace}
\begin{tabular}{@{}lccc@{}}
\toprule
Backbone & 2-pin & Key & Avg. \\
\midrule
Vision-only DP & 6/30 & 5/30 & 18.3\% \\
DP + PACE & 28/30 & 25/30 & 88.3\% \\
\bottomrule
\end{tabular}
\end{wraptable}

\noindent
As shown in \linkto{tab:dp_pace}{Table~4}, DP+PACE reaches 88.3\% average success on 2-pin and Key, a 70.0 pp improvement over vanilla DP on the same two tasks (\linkto{tab:real_results}{Table~1(a)}). These results indicate that PACE's temporal-role organization remains effective across different policy backbones. \appref{app:dp_pace} provides full implementation details.

\WFclear
\subsection{Demonstration Quality from Force Feedback}\label{sec:ff_quality}

This experiment examines whether force feedback improves demonstration quality. The main comparisons in \linkto{tab:real_results}{Tables~1}--\linkto{tab:ablations}{2} use force-feedback demonstrations; here we fix the sigma.7 device, operator, mapping, workspace, and task setup. Within each feedback condition, PACE and ACT + Tactile Concat use identical 2-pin demonstrations; only force-feedback rendering differs between conditions. We use 2-pin because operators achieved below 30\% success on the other tasks without force feedback, making it impractical to collect sufficient successful trajectories.

Under this protocol, feedback-off training reduces success from 29/30 to 21/30 for PACE and from 17/30 to 7/30 for ACT + Tactile Concat (\linkto{tab:ff_policy_success}{Table~\ref{tab:ff_policy_success}}). Feedback-off demonstrations also increase force variance from 4.8 to 12.7~N\(^2\). \appref{app:ff_demo_quality} and \figref{fig:ff_full} provide full trajectory-level evidence: feedback-on demonstrations show tighter force bundles, lower inter-trajectory deviation (1.07 vs.\ 1.92~N), and lower peak forces (10.0$\pm$2.2 vs.\ 16.7$\pm$7.1~N). Thus, force feedback narrows the behavioral distribution, while PACE retains a higher success rate under feedback-off collection in this controlled 2-pin setting.

\section{Conclusion and Limitations}\label{sec:conclusion}

PACE improves chunk-based insertion by organizing tactile feedback into contact-state conditioning, contact-event correction, and contact-consequence prediction rather than single-path fusion. Across real tasks, ManipulationNet, ablations, alternative feedback signals, and DP integration, this design outperforms vision-only and generation-time fusion baselines while preserving chunk continuity.

\textbf{Limitations:} Validation focuses on precision insertion; tasks with different contact-phase structures remain untested. Event correction depends on physical-feedback quality and can fail when coupled lift-translation accumulates object tilt under weak grasping (\appref{app:failure}). Future work should test broader contact tasks, adaptive time scales, and inference-time use of predicted contact consequences.

\bibliography{PACE_camera_ready_v5_refs}

\clearpage
\begin{center}
{\LARGE\bfseries PACE: Temporal-Role Tactile Conditioning for Chunk-Based Insertion Policies}\\[1.5em]
{\Large\bfseries Appendix}
\end{center}
\appendix
\setcounter{figure}{0}
\makeatletter
\@addtoreset{figure}{section}
\@addtoreset{table}{section}
\makeatother
\renewcommand{\thefigure}{\thesection\arabic{figure}}
\setcounter{table}{0}
\renewcommand{\thetable}{\thesection\arabic{table}}
\section{Network Architecture}\label{app:network}

The main text describes PACE according to temporal roles. This appendix gives the concrete parameterization used in the current implementation: contact-state conditioning \(\phi_\text{phase}\), event gate \(\phi_\text{gate}\), event residual \(\phi_\text{res}\), and contact-consequence prediction head \(\phi_\text{pred}\).

\subsection{ACT Backbone}\label{app:act_backbone}

PACE is built on the ACT CVAE-Transformer architecture \citep{ref14}, with layer normalization \citep{ref34} in each layer. During training, the CVAE encoder maps expert action sequences and joint states to a latent variable \(\mathbf{z} \in \mathbb{R}^{32}\). During inference, the policy generates an action chunk of length \(H\) from the current vision, proprioception, and tactile conditions.

\begin{table}[!htbp]
\centering
\caption{ACT backbone hyperparameters.}
\label{tab:act_backbone}
\begin{tabular}{@{}lr@{}}
\toprule
Parameter & Value\tabularnewline
\midrule
Transformer encoder layers & 4\tabularnewline
Transformer decoder layers & 7\tabularnewline
Hidden dimension \(d_{\text{model}}\) & 512\tabularnewline
Attention heads & 8\tabularnewline
Feed-forward dimension & 3200\tabularnewline
CVAE latent dimension & 32\tabularnewline
Action dimension \(D_a\) & 7\tabularnewline
Chunk size \(H\) & 20\tabularnewline
\bottomrule
\end{tabular}
\end{table}

Global and wrist RGB images are processed by the vision encoder and projected to the Transformer hidden dimension. The proprioceptive state \(\mathbf{q}_t\) is encoded as a proprioceptive token. The complete generation pipeline is:

\begin{equation}
\mathbf{h}_s = \text{Encode}(\mathbf{I}^g_s, \mathbf{I}^w_s, \mathbf{q}_s), \quad \mathbf{h}'_s = \phi_{\text{phase}}(\mathbf{h}_s, \mathbf{w}^{\text{state}}_s), \quad \mathbf{A}^{\text{gen}}_s = \text{Decode}(\mathbf{h}'_s, \mathbf{z}),
\label{eq:generation_pipeline}
\end{equation}

where \(\mathbf{I}^g_s\) and \(\mathbf{I}^w_s\) are the global and wrist RGB images, \(\mathbf{q}_s\) is the proprioceptive state, \(\mathbf{h}_s\) is the encoded observation, \(\mathbf{h}'_s\) is the phase-modulated feature, \(\mathbf{z}\) is the CVAE latent variable (sampled from the prior during inference), and \(\mathbf{A}^{\text{gen}}_s\) is the generated action chunk. Contact-state conditioning \(\phi_{\text{phase}}\) modulates encoder features before decoding (\secref{sec:phase}); after chunk generation, contact-event correction further adjusts the generated chunk (\secref{sec:event}).

\subsection{Contact-State Conditioning}\label{app:phase_arch}

The contact state is computed as defined in \secref{sec:phase} (default \(\alpha=0.05\)). The state encoder maps \(\mathbf{w}^{\text{state}}_s\) to a 128-dimensional conditioning vector using a two-layer multi-layer perceptron (MLP) with hidden dimension 128 and ReLU activation. Each encoder layer uses an independent linear modulator, which maps the conditioning vector to \(2 \times d_{\text{model}}\) dimensions and splits it into \((\gamma^l_s, \beta^l_s) \in \mathbb{R}^{512} \times \mathbb{R}^{512}\). These parameters are applied to the encoder activations after LayerNorm, as shown in \eqnref{eq:phase_affine} in \secref{sec:phase}. Independent modulation parameters for each layer allow different layers to respond to contact-phase changes at different granularities.

\subsection{Contact-Event Correction of the Generated Chunk}\label{app:event_arch}

Event correction is applied after action-chunk generation and recomputed at each control step using the latest tactile history. The event cue is constructed as defined in \secref{sec:event} (default \(h_{\text{corr}}=5\), corresponding to 0.25 s under a 20 Hz control rate). We chose \(h_{\text{corr}}=5\) as a balance between responsiveness and stability; in preliminary tests, values in {[}3, 7{]} yielded similar performance.

\begin{table}[!htbp]
\centering
\caption{Event correction module parameterization.}
\label{tab:event_modules}
\begin{tabular}{@{}lllll@{}}
\toprule
\begin{minipage}[b]{0.17\columnwidth}\raggedright
Component\strut
\end{minipage} & \begin{minipage}[b]{0.17\columnwidth}\raggedright
Input\strut
\end{minipage} & \begin{minipage}[b]{0.17\columnwidth}\raggedright
Hidden layers\strut
\end{minipage} & \begin{minipage}[b]{0.17\columnwidth}\raggedright
Output\strut
\end{minipage} & \begin{minipage}[b]{0.17\columnwidth}\raggedright
Final activation\strut
\end{minipage}\tabularnewline
\midrule
\begin{minipage}[t]{0.17\columnwidth}\raggedright
\(\phi_{\text{gate}}\)\strut
\end{minipage} & \begin{minipage}[t]{0.17\columnwidth}\raggedright
\(\mathbf{e}_t \in \mathbb{R}^{D_w+1}\)\strut
\end{minipage} & \begin{minipage}[t]{0.17\columnwidth}\raggedright
64, 64\strut
\end{minipage} & \begin{minipage}[t]{0.17\columnwidth}\raggedright
\(D_a\)\strut
\end{minipage} & \begin{minipage}[t]{0.17\columnwidth}\raggedright
sigmoid\strut
\end{minipage}\tabularnewline
\begin{minipage}[t]{0.17\columnwidth}\raggedright
\(\phi_{\text{res}}\)\strut
\end{minipage} & \begin{minipage}[t]{0.17\columnwidth}\raggedright
\(\text{flatten}(\mathbf{W}_t) \in \mathbb{R}^{T_w \times D_w}\)\strut
\end{minipage} & \begin{minipage}[t]{0.17\columnwidth}\raggedright
128, 128\strut
\end{minipage} & \begin{minipage}[t]{0.17\columnwidth}\raggedright
\(h_{\text{corr}} \times D_a\)\strut
\end{minipage} & \begin{minipage}[t]{0.17\columnwidth}\raggedright
-\strut
\end{minipage}\tabularnewline
\bottomrule
\end{tabular}
\end{table}

For each policy query, Event applies a gated residual to the first \(h_{\text{corr}}\) actions of the generated chunk. The corrected chunk is then passed to the execution mechanism.

\subsection{Contact-Consequence Prediction}\label{app:pred_arch}

The prediction head follows the definition in \secref{sec:pred} and maps decoder hidden states to future tactile changes.

\begin{table}[!htbp]
\centering
\caption{Contact-consequence prediction head parameterization.}
\label{tab:pred_head}
\begin{tabular}{@{}llll@{}}
\toprule
\begin{minipage}[b]{0.22\columnwidth}\raggedright
Component\strut
\end{minipage} & \begin{minipage}[b]{0.22\columnwidth}\raggedright
Input\strut
\end{minipage} & \begin{minipage}[b]{0.22\columnwidth}\raggedright
Hidden layers\strut
\end{minipage} & \begin{minipage}[b]{0.22\columnwidth}\raggedright
Output\strut
\end{minipage}\tabularnewline
\midrule
\begin{minipage}[t]{0.22\columnwidth}\raggedright
\(\phi_{\text{pred}}\)\strut
\end{minipage} & \begin{minipage}[t]{0.22\columnwidth}\raggedright
\(\mathbf{z}_s \in \mathbb{R}^{d_{\text{model}}}\)\strut
\end{minipage} & \begin{minipage}[t]{0.22\columnwidth}\raggedright
512 (ReLU)\strut
\end{minipage} & \begin{minipage}[t]{0.22\columnwidth}\raggedright
\(D_w\)\strut
\end{minipage}\tabularnewline
\bottomrule
\end{tabular}
\end{table}

This head is used only during training and removed during inference.

\subsection{Parameter Count}\label{app:param_count}

\begin{table}[!htbp]
\centering
\caption{Additional parameter count of PACE modules.}
\label{tab:param_count}
\begin{tabular}{@{}lr@{}}
\toprule
Component & Additional Parameters\tabularnewline
\midrule
\(\phi_{\text{phase}}\) (condition encoder + modulator) & 0.546 M\tabularnewline
\(\phi_{\text{gate}}\) & 0.005 M\tabularnewline
\(\phi_{\text{res}}\) & 0.024 M\tabularnewline
\(\phi_{\text{pred}}\) (training only) & 0.266 M\tabularnewline
\textbf{PACE total} & \textbf{0.841 M}\tabularnewline
\bottomrule
\end{tabular}
\end{table}

The ACT backbone has about 83.87 M parameters. PACE adds about 0.841 M parameters, about 1.0\% of the backbone size.

\subsection{Inference Procedure}\label{app:inference}

At each control step \(t\), the inference procedure is:

\begin{enumerate}
\def\labelenumi{\arabic{enumi}.}
\item
  Read the current observation \(\mathbf{o}_t\) (RGB, proprioception, tactile).
\item
  Update the contact state: \(\mathbf{w}^{\text{state}}_t \leftarrow (1{-}\alpha)\mathbf{w}^{\text{state}}_{t-1} + \alpha\mathbf{w}_t\).
\item
  Run the full PACE forward pass: encode visual and proprioceptive features, apply Phase modulation with \(\mathbf{w}^{\text{state}}_t\), decode an action chunk \(\mathbf{A}^{\text{gen}}_t\), and apply Event correction to the front \(h_{\text{corr}}\) steps using the latest tactile history \(\mathbf{W}_t\).
\item
  Store the corrected chunk in the temporal-ensemble buffer.
\item
  Aggregate overlapping chunk predictions at time \(t\) using exponential-decay weights to obtain the final action \(\mathbf{a}^{\text{exec}}_t\).
\item
  Dispatch \(\mathbf{a}^{\text{exec}}_t\) to the robot.
\end{enumerate}

Steps 1--6 repeat at each control step. The corrected chunk is then consumed by the standard execution mechanism of the backbone policy.

\section{Force-Feedback Data-Collection System}\label{app:collection}

\subsection{System Overview}\label{app:system_overview}

Demonstrations are collected using our force-feedback multimodal data-collection system. The system consists of a Realman RM75 robot arm, a Robotiq Hand-E parallel gripper, an end-effector six-axis force/torque sensor, a Vitai GF225 optical visuotactile sensor, two Intel RealSense D405 RGB cameras, and a Force Dimension sigma.7 force-feedback teleoperation device. A sigma.7 background process reads the operator's hand pose at high frequency and outputs force feedback, while the main collection process performs robot control, sensor reading, and HDF5 writing at 20 Hz.

The goal of the system is to make the operator ``feel what they control'' in the contact-point coordinate frame: the operator's hand offset is interpreted as a control offset in the contact-point frame and transformed to drive the robot; the contact force measured by the robot end-effector sensor is also transformed back to the same contact-point frame and rendered to the operator.

\subsection{Contact-Point Anchoring}\label{app:anchoring}

Before each trajectory, the robot moves to a reset pose and the sigma.7 is centered. When the rising edge of the teleoperation enable signal is detected, the system synchronously records:

\begin{equation}
\mathbf{x}^{r}_0 = \text{robot anchor pose}, \quad
\mathbf{x}^{d}_0 = \text{device anchor pose}.
\label{eq:anchor_pose}
\end{equation}

Subsequent device poses are represented as offsets relative to the anchor. If the operator pauses collection and re-enables teleoperation, the system rebuilds the anchor and continues collection from the current robot pose. This mechanism reduces pose jumps at startup.

\subsection{Action Mapping}\label{app:action_mapping}

Let \({E}\) be the end-effector coordinate frame and \({T}\) be the contact-point coordinate frame. \(R_{ET}\) maps vectors from \({T}\) to \({E}\), and \(p_{ET}\) is the displacement from the end-effector origin to the contact point, expressed in \({E}\).

The device-side position offset \(\Delta p_T\) is mapped to the end-effector frame by

\begin{equation}
\Delta p_E = R_{ET}\Delta p_T.
\label{eq:delta_pos}
\end{equation}

If the device-side rotation offset is \(R_{\Delta T}\), the corresponding end-effector rotation increment is

\begin{equation}
R_{\Delta E}=R_{ET}R_{\Delta T}R_{ET}^{T}.
\label{eq:delta_rot}
\end{equation}

The target end-effector pose is

\begin{equation}
p_{\text{target}}=p_0+\Delta p_E, \quad
R_{\text{target}}=R_{\Delta E}R_0.
\label{eq:target_pose}
\end{equation}

\subsection{Wrench Transformation for Force Feedback}\label{app:wrench_transform}

The force-feedback loop follows a pipeline of low-pass filtering, bias compensation, wrench transformation, contact-point semantic-space shaping, and device output.

First, the bias is subtracted in the end-effector frame to obtain the pure contact wrench:
\begin{equation}
F_E = \widetilde{F}_E - b_F, \quad
\tau_E = \widetilde{\tau}_E - b_\tau.
\label{eq:bias_subtraction}
\end{equation}

The force and torque in the end-effector frame are then transformed to the contact-point frame:

\begin{equation}
F_T = R_{ET}^{T}F_E,
\label{eq:force_transform}
\end{equation}

\begin{equation}
\tau_T = R_{ET}^{T}(\tau_E - p_{ET}\times F_E).
\label{eq:torque_transform}
\end{equation}

Here, \(p_{ET}\times F_E\) denotes the equivalent torque term induced by the moment arm from the end-effector sensor origin to the contact point. In matrix form:

\begin{equation}
\begin{bmatrix}
F_T \\
\tau_T
\end{bmatrix}
=
\begin{bmatrix}
R_{ET}^{T} & 0 \\
-[R_{ET}^{T}p_{ET}]_\times R_{ET}^{T} & R_{ET}^{T}
\end{bmatrix}
\begin{bmatrix}
F_E \\
\tau_E
\end{bmatrix}.
\label{eq:wrench_matrix}
\end{equation}

Force and torque are independently scaled in the contact-point frame:

\begin{equation}
F_T^{\text{out}}=k_F F_T, \quad
\tau_T^{\text{out}}=k_\tau \tau_T.
\label{eq:force_torque_scaling}
\end{equation}

This scaling is performed in the contact-point frame because each component has a clear task meaning: axial insertion resistance, lateral collision force, and torsional misalignment torque. Finally, the system outputs the scaled wrench to the force-feedback device according to the sigma.7 coordinate mapping.

\subsection{Collection Gating and Data Format}\label{app:gating_format}

The collection process uses an enable gate and a warm-up discard mechanism. When teleoperation is not enabled, the system does not advance the valid step count or write samples. After enable, the first \(K\) control cycles are not written, suppressing startup transients. Each valid step writes multimodal observations and actions:

\begin{verbatim}
dataset.hdf5
|-- observations/
|   |-- images/
|   |   |-- global              [T, 200, 266, 3]   global RGB
|   |   |-- wrist               [T, 200, 266, 3]   wrist RGB
|   |-- proprio_joint           [T, 7]              joint state
|   |-- proprio_eef             [T, 6]              end-effector pose
|   |-- tac/
|       |-- left/
|       |   |-- force6d         [T, 6]              GF225 tactile force
|       |   |-- img             [T, 240, 240, 3]    tactile image
|       |   |-- depth           [T, 240, 240]       tactile depth
|       |   |-- marker_offset   [T, 9, 9, 2]        marker displacement
|       |-- right/              (same structure)
|-- ft                          [T, 6]              end-effector wrench
|-- joint_current               [T, 7]              joint current
|-- actions/
    |-- joint_abs               [T, 7]              target joint action
    |-- eef_abs                 [T, 6]              absolute EE action
\end{verbatim}

\section{Training Details}\label{app:training}

\subsection{Hyperparameters}\label{app:hyperparameters}

\begin{table}[!htbp]
\centering
\caption{Training hyperparameters.}
\label{tab:training_hyperparameters}
\begin{tabular}{@{}lr@{}}
\toprule
Parameter & Value\tabularnewline
\midrule
Optimizer & AdamW\tabularnewline
Learning rate & \(1\times10^{-5}\)\tabularnewline
Batch size & 32\tabularnewline
Training epochs & 10000\tabularnewline
Chunk size \(H\) & 20\tabularnewline
KL weight \(\lambda_{\text{kl}}\) & 10\tabularnewline
Prediction weight \(\lambda_{\text{pred}}\) & 0.05\tabularnewline
Smoothing rate \(\alpha\) & 0.05\tabularnewline
Tactile history \(T_w\) & 4\tabularnewline
Correction horizon \(h_{\text{corr}}\) & 5\tabularnewline
Training hardware & 1\(\times\) RTX 3090 (24 GB)\tabularnewline
\bottomrule
\end{tabular}
\end{table}

\subsection{Training Convergence}\label{app:training_convergence}

All models are trained for 10000 epochs. We select the checkpoint with the lowest validation loss for each task configuration. Across all 11 configurations (5 real tasks + 6 ManipulationNet geometry--tolerance settings), the best checkpoints fall between epochs 7102 and 8728:

\begin{table}[!htbp]
\centering
\caption{Selected checkpoints by task.}
\label{tab:training_convergence}
\begin{tabular}{@{}lr@{}}
\toprule
Task & Best epoch\tabularnewline
\midrule
Key & 7456\tabularnewline
2-pin & 7102\tabularnewline
3-pin & 8126\tabularnewline
USB & 8003\tabularnewline
Ethernet & 7914\tabularnewline
Cyl-3mm & 7571\tabularnewline
Cyl-1mm & 7419\tabularnewline
Cyl-0.1mm & 7356\tabularnewline
Hex-3mm & 8728\tabularnewline
Hex-1mm & 7130\tabularnewline
Hex-0.1mm & 7122\tabularnewline
\bottomrule
\end{tabular}
\end{table}

No early stopping is used; instead, we evaluate all saved checkpoints on a held-out validation set and deploy the one with the lowest validation action loss. The convergence range (7100--8800 epochs) is consistent across tasks, indicating stable training dynamics.

\subsection{Loss}\label{app:loss}

PACE uses the joint loss described in \secref{sec:objective}. The action loss \(\mathcal{L}_{\text{act}}\) supervises the action chunk after Event correction, where the first \(h_{\text{corr}}\) steps are Event-corrected actions and the remaining steps are uncorrected generated actions. \(\mathcal{L}_{\text{pred}}\) is the mean squared error (MSE) loss for future tactile changes, and padding positions near the end of an episode are masked.

\subsection{Signal Preprocessing}\label{app:signal_preprocessing}

Images are normalized to \([0,1]\) and standardized using ImageNet statistics. Proprioceptive states, actions, and low-dimensional tactile signals are z-score normalized using training-set statistics. When the tactile history is shorter than \(T_w\) frames, the most recent available frame is used for padding.

\subsection{Dataset Statistics}\label{app:dataset_statistics}

\begin{table}[!htbp]
\centering
\caption{Dataset and observation statistics.}
\label{tab:dataset_statistics}
\begin{tabular}{@{}ll@{}}
\toprule
Item & Value\tabularnewline
\midrule
Demonstrations per task & \textasciitilde160 episodes\tabularnewline
Episode length & \textasciitilde300 steps\tabularnewline
RGB image resolution & 200 \(\times\) 266\tabularnewline
Tactile sensor resolution (Vitai GF225) & 240 \(\times\) 240\tabularnewline
Action space & 7-DoF joint position\tabularnewline
\bottomrule
\end{tabular}
\end{table}

\subsection{Evaluation Protocol and Task Definitions}\label{app:evaluation_protocol}

The real-robot evaluation is designed to test contact-rich insertion under moderate pose randomization rather than repeated execution from a single fixed pose. For each method and each task, we run 30 independent rollouts. Before each rollout, the target socket is randomized by moving the whiteboard that holds the socket within a 25 mm \(\times\) 25 mm range. This range is large enough to change the contact condition and local search trajectory, while remaining within the demonstrated workspace.

Before each rollout, we reset the robot and zero the force/tactile sensors. An automatic force-threshold safety stop is triggered by excessive wrist six-axis force; such trials count as failures. Per-step action clamping is implemented in the deployment client but was disabled in these comparisons to measure the policy's native outputs.

A rollout is counted as successful only if the connector or key is inserted within the time limit and the final state is mechanically stable. Stability means that the inserted object either maintains contact without immediate detachment or reaches a functional connection state, depending on the object. Failures include no contact, edge jamming, excessive contact force that triggers safety stopping, timeout, and post-insertion detachment.

The five real tasks were selected to cover distinct contact ambiguities:

\begin{enumerate}
\def\labelenumi{\arabic{enumi}.}
\item
  \textbf{Key insertion.} The slot is non-axisymmetric and imposes a clear orientation constraint. Small rotational errors cause side interference, rim contact, or jamming. This task stresses the ability to distinguish productive contact from contact that requires retreat or re-alignment.
\item
  \textbf{Two-pin plug insertion.} The dual-pin geometry creates a parallel constraint. During approach, the wrist camera does not directly observe the pins, so the policy must infer contact state and correct local misalignment after first contact.
\item
  \textbf{Three-pin plug insertion.} This task adds asymmetric multi-point constraints and tighter pose tolerance. It is difficult for vision-only policies because neither the global nor wrist view reliably exposes the pin-hole relationship near contact.
\item
  \textbf{USB insertion.} USB insertion has small clearance and strong pose sensitivity. A small translational or rotational error can lead to rim collision, requiring local search before successful insertion.
\item
  \textbf{Ethernet insertion.} Ethernet contains an elastic latch and requires sustained insertion force. Here a force increase can indicate either normal progress or harmful jamming, making force semantics especially important.
\end{enumerate}

These tasks jointly test whether a policy can handle both slowly varying contact phases, such as search and stable insertion, and abrupt contact events, such as collision or jamming.

\subsection{Baseline Implementation Details}\label{app:baseline_implementation}

All baselines use the same demonstrations, image streams, action space, control frequency, and evaluation protocol as PACE. This keeps the comparison focused on policy architecture and contact-feedback usage.

\textbf{Vision-only ACT.} The ACT baseline follows the same CVAE-Transformer backbone as PACE but removes all tactile inputs and PACE-specific modules. It predicts action chunks from RGB and proprioception only. This baseline tests whether chunk-based visual imitation alone can solve the insertion tasks.

\textbf{Vision-only Diffusion Policy.} The DP baseline uses a diffusion model to represent the action distribution from demonstrations. It also receives only RGB and proprioception. This baseline tests whether a more expressive action-generation model can compensate for the lack of contact feedback.

\textbf{\(\pi_0\).} The \(\pi_0\) baseline is fine-tuned on the same demonstrations with the same visual inputs and action space as PACE. It tests whether large-scale vision-language-action pretraining can overcome post-contact ambiguity without tactile feedback.

\textbf{ACT + Tactile Cross-Attn.} This baseline adds a cross-attention layer where visual encoder tokens attend to tactile tokens projected from the same tactile input. It represents attention-based visuotactile fusion at generation time.

\textbf{ACT + Tactile Concat.} This baseline projects the tactile signal into a token and concatenates it with the encoder input sequence. It represents a single-path generation-time tactile-fusion design.

\subsection{ManipulationNet Evaluation Details}\label{app:manip_eval}

ManipulationNet provides a standardized NIST Assembly Task Board peg-in-hole benchmark with cylindrical and hexagonal pegs. Each peg geometry is paired with holes at three clearance levels: 3 mm, 1 mm, and 0.1 mm. The 0.1 mm setting is especially challenging because visual pose error, small fixture uncertainty, and local rim contact can dominate the insertion outcome.

PACE is evaluated on all six geometry-clearance configurations. Each configuration uses 25 independent rollouts. \linkto{tab:manip_pace}{Table~1(b)} reports PACE disaggregated by geometry and clearance, while \appref{app:manip_baselines} reports baseline performance averaged across peg geometries. The ManipulationNet results are used as a standardized complement to the five real connector tasks: the benchmark isolates tolerance and peg geometry, while the real tasks include practical connector-specific issues such as visual missingness, asymmetric constraints, elastic latches, and post-insertion stability.

\section{Supplementary Analysis}\label{app:supplementary_analysis}

\subsection{Temporal Ensemble and Event Adaptation}\label{app:temporal_ensemble}

The temporal ensemble (TE) in ACT uses exponential-decay weighted averaging over multiple overlapping chunks that predict the same time step, smoothing transitions between chunks. Event and TE act at different levels: TE mixes existing predictions, injects no new information, and uses weights unrelated to contact; Event generates an additive residual from tactile differences and directly modifies action content. They are complementary: TE handles inter-chunk smoothness, while Event handles intra-chunk contact adaptation.

\begin{table}[!htbp]
\centering
\caption{Comparison between temporal ensemble and Event correction.}
\label{tab:te_event_comparison}
\begin{tabular}{@{}lll@{}}
\toprule
\begin{minipage}[b]{0.30\columnwidth}\raggedright
\strut
\end{minipage} & \begin{minipage}[b]{0.30\columnwidth}\raggedright
Temporal Ensemble\strut
\end{minipage} & \begin{minipage}[b]{0.30\columnwidth}\raggedright
Event correction\strut
\end{minipage}\tabularnewline
\midrule
\begin{minipage}[t]{0.30\columnwidth}\raggedright
Operates on\strut
\end{minipage} & \begin{minipage}[t]{0.30\columnwidth}\raggedright
Predictions from multiple overlapping action chunks at the same time step\strut
\end{minipage} & \begin{minipage}[t]{0.30\columnwidth}\raggedright
Generated action chunk\strut
\end{minipage}\tabularnewline
\begin{minipage}[t]{0.30\columnwidth}\raggedright
Mechanism\strut
\end{minipage} & \begin{minipage}[t]{0.30\columnwidth}\raggedright
Fixed exponential weighted averaging\strut
\end{minipage} & \begin{minipage}[t]{0.30\columnwidth}\raggedright
Learned gated residual from latest tactile history\strut
\end{minipage}\tabularnewline
\begin{minipage}[t]{0.30\columnwidth}\raggedright
Signal driven\strut
\end{minipage} & \begin{minipage}[t]{0.30\columnwidth}\raggedright
No\strut
\end{minipage} & \begin{minipage}[t]{0.30\columnwidth}\raggedright
Yes, tactile differential\strut
\end{minipage}\tabularnewline
\begin{minipage}[t]{0.30\columnwidth}\raggedright
Contact-aware\strut
\end{minipage} & \begin{minipage}[t]{0.30\columnwidth}\raggedright
No\strut
\end{minipage} & \begin{minipage}[t]{0.30\columnwidth}\raggedright
Yes\strut
\end{minipage}\tabularnewline
\bottomrule
\end{tabular}
\end{table}

\subsection{Action-Chunk Length Analysis}\label{app:chunk_length}

The short-chunk control uses Phase-only and shortens the chunk length from \(H=20\) to \(H=10\). This experiment asks whether simply increasing the query frequency is sufficient to replace Event's contact-change-driven correction, when the model already uses tactile feedback during generation.

\begin{table}[!htbp]
\centering
\caption{Short-chunk control on real insertion tasks.}
\label{tab:chunk_length_appendix}
\begin{tabular}{@{}lccccccc@{}}
\toprule
Method & Chunk & Key & 2-pin & 3-pin & USB & Eth. & Avg.\tabularnewline
\midrule
Phase-only & 20 & 17/30 & 20/30 & 16/30 & 18/30 & 19/30 & 60.0\%\tabularnewline
Phase-only (chunk=10) & 10 & 14/30 & 17/30 & 13/30 & 15/30 & 15/30 & 49.3\%\tabularnewline
PACE & 20 & 27/30 & 29/30 & 28/30 & 27/30 & 25/30 & 90.7\%\tabularnewline
\bottomrule
\end{tabular}
\end{table}

Shortening the chunk from 20 to 10 reduces success from 60.0\% to 49.3\%, with the largest drop on Ethernet (19/30 → 15/30), which requires sustained insertion force through an elastic latch. This suggests that higher query frequency cannot substitute for temporal-role organization: shorter chunks weaken temporal coherence and sustained-force behavior.

\subsection{Comparison with RDP}\label{app:rdp}

RDP uses the authors' public implementation; both methods use the same demonstrations and evaluation protocol.

\begin{table}[!htbp]
\centering
\caption{Comparison with RDP on two real insertion tasks (successes/30).}
\label{tab:rdp}
\begin{tabular}{@{}lcc@{}}
\toprule
Method & 2-pin & Key\tabularnewline
\midrule
RDP & 19/30 & 12/30\tabularnewline
\textbf{PACE} & \textbf{29/30} & \textbf{27/30}\tabularnewline
\bottomrule
\end{tabular}
\end{table}

\subsection{Statistical Reliability}\label{app:statistical_reliability}

For the 30-rollout real-robot comparisons, we report Wilson 95\% confidence intervals computed from the observed success and failure counts.

\begin{table}[!htbp]
\centering
\caption{Real-robot successes and Wilson 95\% confidence intervals.}
\label{tab:real_ci}
\begin{tabular}{@{}lll@{}}
\toprule
Task & PACE & ACT + Tactile Concat\tabularnewline
\midrule
Key & 27/30 [0.74, 0.97] & 13/30 [0.27, 0.61]\tabularnewline
2-pin & 29/30 [0.83, 0.99] & 17/30 [0.39, 0.73]\tabularnewline
3-pin & 28/30 [0.79, 0.98] & 12/30 [0.25, 0.58]\tabularnewline
USB & 27/30 [0.74, 0.97] & 13/30 [0.27, 0.61]\tabularnewline
Ethernet & 25/30 [0.66, 0.93] & 14/30 [0.30, 0.64]\tabularnewline
\bottomrule
\end{tabular}
\par\smallskip

\emph{Entries are successes/30 followed by Wilson 95\% confidence intervals.}
\end{table}

\subsection{ManipulationNet Baseline Performance}\label{app:manip_baselines}

To contextualize PACE's ManipulationNet results (\linkto{tab:manip_pace}{Table~1(b)}), we report baseline success rates on the same benchmark across clearance levels. All baselines are trained and evaluated by us under identical conditions. Each configuration is evaluated with 25 independent rollouts using the same evaluation protocol as PACE.

\begin{table}[!htbp]
\centering
\caption{Baseline success rates on ManipulationNet, averaged over cylindrical and hexagonal pegs. Each geometry-clearance configuration uses 25 rollouts.}
\label{tab:manip_baselines}
\begin{tabular}{@{}lccc@{}}
\toprule
Method & 3 mm & 1 mm & 0.1 mm\tabularnewline
\midrule
Vision-only ACT & 30\% & 10\% & 0\%\tabularnewline
Vision-only DP & 10\% & 0\% & 0\%\tabularnewline
\textbf{PACE} & \textbf{98\%} & \textbf{88\%} & \textbf{80\%}\tabularnewline
\bottomrule
\end{tabular}
\end{table}

Vision-only ACT achieves moderate success at the loosest clearance (3 mm, 30\%) but degrades rapidly as tolerance tightens, reaching 0\% at 0.1 mm. Vision-only DP fails entirely below 3 mm clearance. These results are consistent with ReTac-ACT \citep{ref35}, which reports that pure-vision policies struggle with industrial-grade clearances due to visual occlusion during the contact phase. In contrast, PACE maintains 80\% success even at 0.1 mm, showing that the temporal-role tactile decomposition provides robust contact-phase guidance across the full tolerance range.

\subsection{Cross-Architecture Validation: DP+PACE Implementation}\label{app:dp_pace}

We integrate the three temporal roles of PACE into Diffusion Policy \citep{ref15} to test cross-backbone transfer. This section details the architecture, integration, training, and deployment.

\textbf{Base Diffusion Policy architecture.} The backbone is a 1D Conditional UNet with channel dimensions [512, 1024, 2048], kernel size~5, and 8 GroupNorm groups. The vision encoder consists of two independent ResNet18 networks (one per camera, not weight-shared), with ImageNet normalization and random crop augmentation (200$\times$266 input, 180$\times$240 crop). Proprioception (7D joint angles) is concatenated with the flattened vision features to form the observation conditioning vector. The diffusion process uses a DDPM scheduler with 100 training timesteps, squared-cosine beta schedule ($\beta_{\text{start}}{=}10^{-4}$, $\beta_{\text{end}}{=}0.02$), epsilon prediction, and \texttt{clip\_sample=true}. The action chunk has horizon $H{=}20$ with $n\_\text{action\_steps}{=}8$ and $n\_\text{obs\_steps}{=}2$.

\textbf{Contact-state conditioning (Phase).} The default GF225 tactile-force signal (6D) is provided as an additional observation key with an extended history window ($T_w{=}4$ frames, longer than the 2-frame visual history). At each query, the EMA-smoothed tactile state $\mathbf{w}^{\text{state}}$ is computed from this history ($\alpha{=}0.05$) and encoded by a small MLP into a phase conditioning vector. This vector is appended to the UNet's global conditioning:
\begin{equation}
\mathbf{c}_{\text{global}} = [\mathbf{h}_{\text{obs}},\; \phi_{\text{phase}}(\mathbf{w}^{\text{state}}_s)].
\label{eq:dp_global_conditioning}
\end{equation}
The phase embedding influences every residual block of the UNet through FiLM conditioning, modulating the denoising process based on the current contact state. The tactile signal is not fed into the visual encoder, ensuring a fair comparison.

\textbf{Contact-event correction (Event).} After diffusion sampling produces a denoised action chunk $\hat{\mathbf{a}}$, the gated residual corrector modifies the first $h_{\text{corr}}$ steps (auto-set to $H/4{=}5$):
\begin{equation}
\hat{\mathbf{a}}[s{:}s{+}h_{\text{corr}}] \leftarrow \hat{\mathbf{a}}[s{:}s{+}h_{\text{corr}}] + \mathbf{g}_t \odot \mathbf{r}_{t:t+h_{\text{corr}}-1}.
\label{eq:dp_event_correction}
\end{equation}
The gate $\mathbf{g}{\in}[0,1]^{D_a}$ is produced by a small MLP with sigmoid output from the tactile-change feature $[\Delta\mathbf{W}, \|\Delta\mathbf{W}\|]$. The residual $\mathbf{r}{\in}\mathbb{R}^{h_{\text{corr}}\times D_a}$ is produced by a separate MLP from the flattened tactile history. During training, the event correction is applied to the clean action target before noise is added, so gradients propagate through the main diffusion loss.

\textbf{Contact-consequence prediction (Aux).} An MLP head predicts the future $\Delta\mathbf{W}$ trajectory ($H$ steps $\times$ 6D) from the global conditioning features, supervised by masked MSE with $\lambda_{\text{pred}}{=}0.05$.

\textbf{Action normalization.} Actions are 7D absolute joint angles (in degrees, range approximately [$-$4, 89]). We apply per-dimension min-max normalization to [$-$1, 1], which is critical for compatibility with the DDPM \texttt{clip\_sample=true} setting. Observations are normalized separately: joint positions use range normalization, the tactile signal uses Gaussian normalization (zero-mean, unit-variance), and images are scaled from [0, 1] to [$-$1, 1].

\textbf{Hyperparameter alignment.} All PACE-specific hyperparameters --- $\alpha$, $T_w$, $h_{\text{corr}}$, $\lambda_{\text{pred}}$, and gate/residual network dimensions --- are identical between ACT+PACE and DP+PACE, isolating the effect of the policy backbone.

\textbf{Training.} The dataset is identical to the ACT experiments. DP+PACE is trained for 8000 epochs with AdamW (lr=$10^{-4}$, betas=(0.95, 0.999), weight decay=$10^{-6}$), a cosine learning rate schedule with 500 warmup steps, and EMA model averaging. Batch size is 64 on a 4090 GPU (or 128 on H20). The model has 310.5M parameters total, of which the PACE-specific modules (Phase encoder + Event gate/residual + Aux head) account for 3.6M (+1.2\%). Training converges at approximately epoch 1100 (680k gradient steps at batch size 64).

\textbf{Deployment.} Following Chi et al.~\citep{ref15}, we train with the DDPM schedule (100 steps) but deploy with DDIM sampling using fewer denoising steps, reducing per-query inference latency to under 10~ms on a 3090 GPU. The policy is served via WebSocket using the same client as ACT, running at 20~Hz. All model parameters and normalization statistics are loaded from the checkpoint.

\subsection{Supplementary Note on Temporal Window Setting}\label{app:temporal_window}

In the main experiments, Phase and Event share \(T_w=4\) to keep the hyperparameter setting simple. In preliminary experiments, using different history windows for Phase and Event, for example a longer window for estimating contact state and a shorter window for constructing event cues, also maintained stable execution performance. This is consistent with PACE's design: Phase captures slowly varying contact context, while Event targets fast-varying contact changes. We do not perform a systematic ablation over window combinations in this paper, and leave learning or selecting different time scales to future work.

We also tested alternative hyperparameter settings: \(\alpha=0.2\) with a Phase history window of 8 steps and an Event window of 4 steps. This configuration also maintained stable execution performance, suggesting that PACE is not sensitive to the specific choice of smoothing rate or window length within a reasonable range.

\subsection{Force-Feedback Demonstration Quality Analysis}\label{app:ff_demo_quality}

This section provides detailed quantitative analysis supporting the force-feedback comparison in \secref{sec:ff_quality}. The feedback-on/off collection comparison fixes the device, operator, mapping, workspace, and task setup; only force-feedback rendering is toggled. Within each feedback condition, PACE and ACT + Tactile Concat use identical 2-pin demonstrations. All metrics are computed on 162 demonstrations per condition (feedback-on vs.\ feedback-off) for the 2-pin insertion task. Force is calibrated by subtracting the first-frame reading, and magnitude is computed as \(\|\mathbf{F}_{\text{cal}}(t)\| = \sqrt{F_x^2 + F_y^2 + F_z^2}\). The trained-policy comparison is restricted to 2-pin because feedback-off demonstrations on Key, 3-pin, USB, and Ethernet had success rates below 30\%, making it impractical to collect enough successful trajectories for controlled policy training. Demonstrations were collected freely without timing instructions.

\begin{table}[htbp]
\centering
\scriptsize
\caption{Policy success under force-feedback collection conditions (2-pin; 30 rollouts per method).}
\label{tab:ff_policy_success}
\begin{tabular}{@{}lcc@{}}
\toprule
Policy & Feedback on & Feedback off \\
\midrule
PACE & 29/30 & 21/30 \\
ACT + Tactile Concat & 17/30 & 7/30 \\
\bottomrule
\end{tabular}
\par\smallskip
\emph{Note:} Within each feedback condition, both policies use identical demonstrations; only force rendering differs between conditions.
\end{table}

\begin{table}[htbp]
\centering
\scriptsize
\caption{Demonstration quality metrics (2-pin, 162 episodes per condition). Relative change is reported as FF-on vs.\ FF-off; where tested, differences are statistically significant (Mann-Whitney U test).}
\label{tab:ff_metrics}
\begin{tabular}{@{}lcccc@{}}
\toprule
Metric & FF on & FF off & Rel. change & $p$-value / $d$ \\
\midrule
Force variance (N\(^2\)) & 4.8 & 12.7 & $-62\%$ & $p<10^{-13}$, $d{=}0.80$ \\
Inter-trajectory deviation (N) & 1.07 & 1.92 & $-44\%$ & $p<10^{-4}$, $d{=}0.55$ \\
Pairwise force-curve correlation & 0.67 & 0.54 & $+24\%$ & --- \\
Peak contact force (N) & $10.0 \pm 2.2$ & $16.7 \pm 7.1$ & $-40\%$ & $p<10^{-24}$, $d{=}1.28$ \\
Episode length CV & 0.108 & 0.132 & $-18\%$ & --- \\
\bottomrule
\end{tabular}
\end{table}

\begin{figure}[htbp]
\centering
\includegraphics[width=\linewidth]{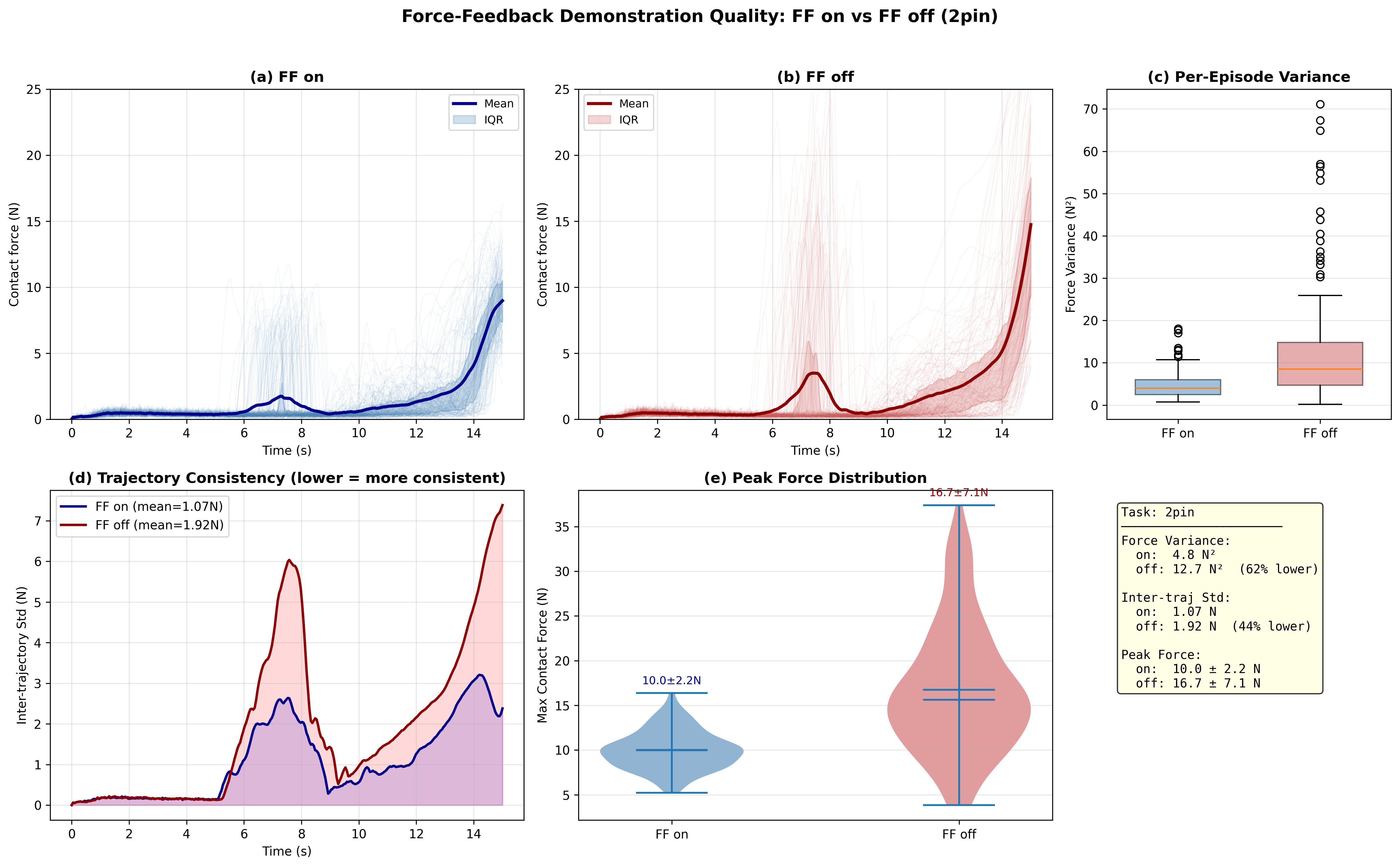}
\caption{Full demonstration quality comparison (2-pin, 162 episodes). (a--b) Force curve overlays. (c) Per-episode variance distribution. (d) Inter-trajectory standard deviation over normalized time. (e) Peak force distribution. Force-feedback demonstrations exhibit consistently lower variance, tighter trajectory bundles, and more predictable peak forces.}
\label{fig:ff_full}
\includegraphics[width=0.7\linewidth]{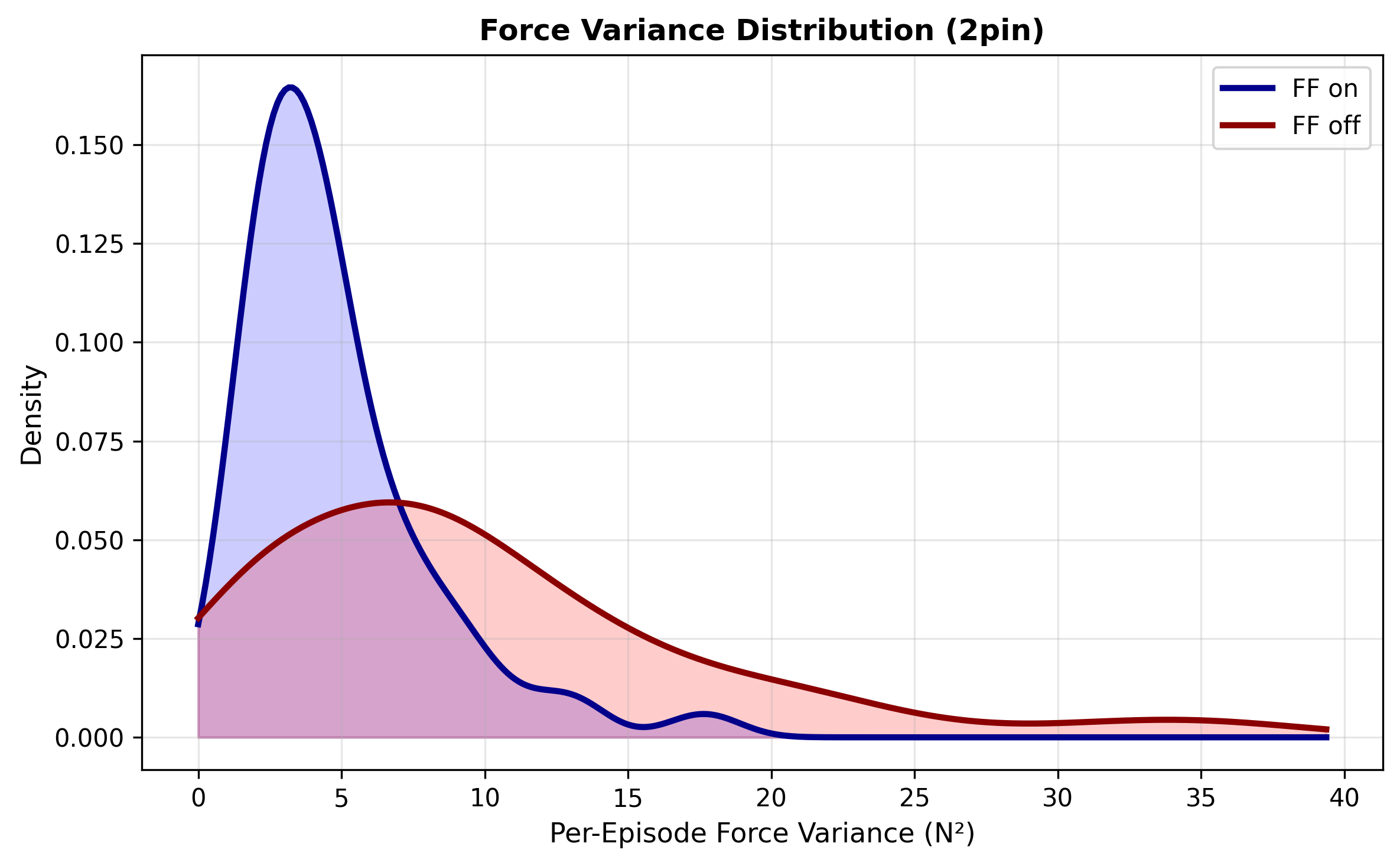}
\captionof{figure}{Per-episode force variance KDE distribution. Feedback-on (blue) concentrates at 3--5~N\(^2\) with rapid density falloff, while feedback-off (red) shows a flat, heavy-tailed distribution extending toward 40~N\(^2\), indicating many episodes with erratic force behavior.}
\label{fig:ff_kde}
\end{figure}

Force feedback enables the operator to regulate contact force in real time, resulting in (1) lower within-trajectory force fluctuation, (2) higher between-trajectory consistency at each task phase, and (3) fewer extreme-force outliers (\figref{fig:ff_kde}). These properties produce a narrower behavioral distribution that is more amenable to imitation learning: the policy faces less ambiguity in what constitutes correct contact behavior at each stage.

\textbf{Task-specific observations.} Force feedback is particularly important for: (1) tasks with visually occluded contact geometry---in 3-pin insertion, the pins become hidden during late-stage alignment, forcing the operator to rely entirely on contact force for alignment feedback; (2) high-precision tasks---at 0.1~mm clearance, subtle lateral forces signal misalignment before jamming, but without force rendering the operator cannot detect these early warnings; (3) depth sensing---across all tasks, force feedback provides the most direct cue for whether insertion is complete or whether excessive force is being applied.

\FloatBarrier
\section{Failure Cases}\label{app:failure}

After collision, the policy produces correction behavior that moves the gripper toward the target. However, when lifting and lateral translation are coupled rather than sequential, and the retreat height is insufficient to fully release surface contact, the peg tip remains on the socket surface and acts as a pivot. Each correction attempt shifts the contact point and accumulates tilt. For narrow, flat objects such as keys, USB connectors, and Ethernet plugs, the weak silicone grip provides minimal resistance to lateral shear, making the object prone to slipping and rotating within the gripper. As shown in \figref{fig:failure_case}, repeated corrections progressively worsen the tilt until the object jams and insertion fails --- despite the gripper consistently moving toward the correct position. This failure can be mitigated by improving demonstration coverage for post-collision recovery trajectories.

\noindent\begin{minipage}{\linewidth}
\centering
\includegraphics[width=\linewidth,totalheight=0.45\textheight,keepaspectratio]{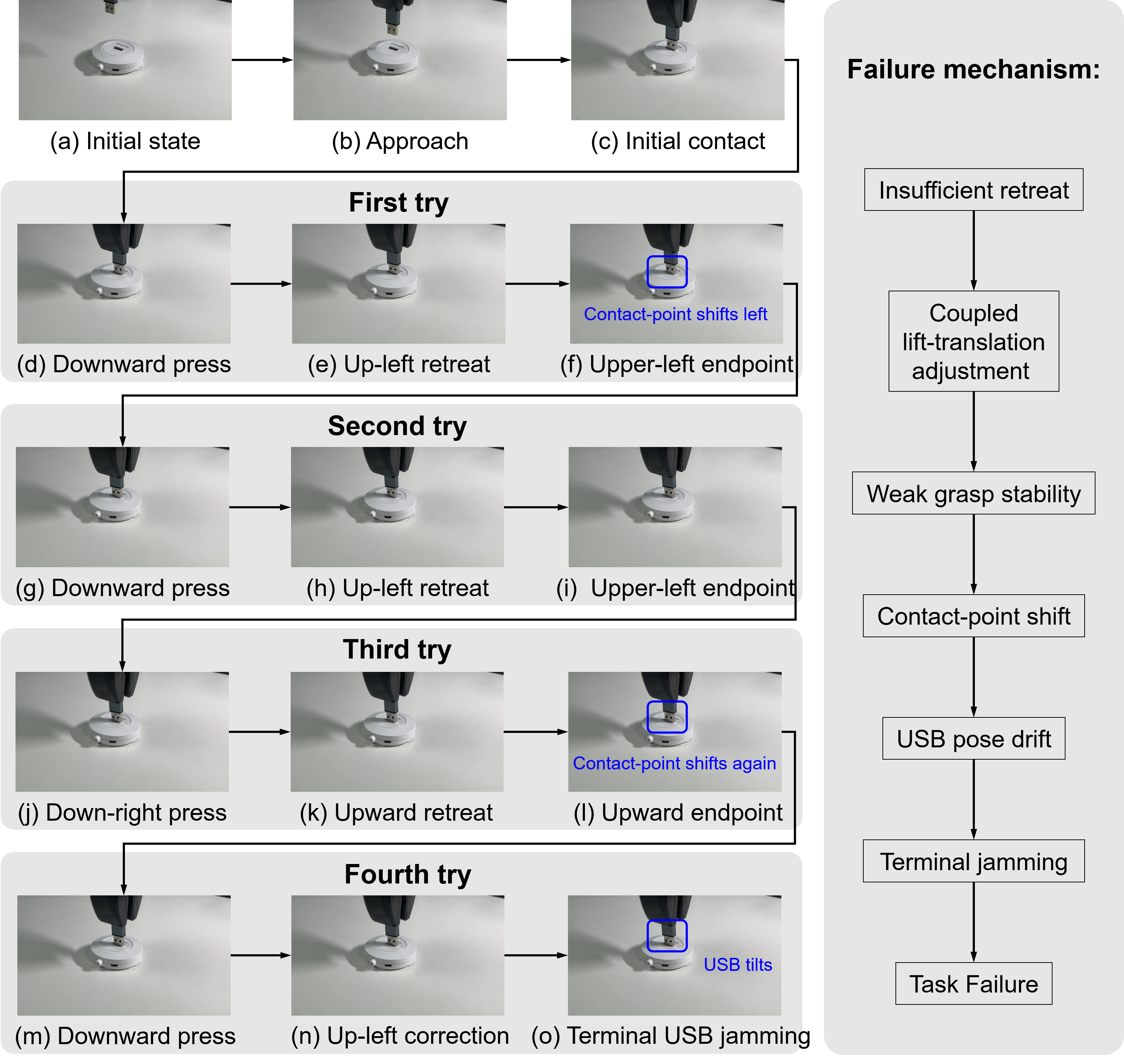}
\captionof{figure}{Representative failure case on USB insertion. The policy produces multiple correction attempts that move toward the target, but coupled lift--translation and insufficient retreat shift the contact point, accumulate object tilt, and lead to terminal jamming.}
\label{fig:failure_case}
\end{minipage}

\section{Computational Overhead}\label{app:overhead}

\begin{table}[!htbp]
\centering
\caption{Computational overhead of PACE modules.}
\label{tab:computational_overhead}
\footnotesize
\begin{tabular}{@{}lrr@{}}
\toprule
Component & Additional Params & Inference Latency\\
\midrule
ACT backbone & 83.87 M & 6.2 ms\\
Contact-state conditioning & 0.546 M & \textless{} 0.1 ms\\
Event gate & 0.005 M & \textless{} 0.1 ms\\
Event residual corrector & 0.024 M & \textless{} 0.1 ms\\
Prediction head & 0.266 M & 0 ms at inference\\
\textbf{PACE training-time total} & \textbf{0.841 M} & --- \\
\textbf{PACE inference-time overhead} & \textbf{0.575 M} & \textbf{\textless{} 0.3 ms}\\
\bottomrule
\end{tabular}
\end{table}

ACT+PACE takes about 6.2 ms per inference step in total, with PACE-specific overhead below 0.3 ms.

\section{Force Calibration}\label{app:force_calibration}

The end effector holds the gripper and task object, so the initial force/torque sensor reading is nonzero. Raw force includes load gravity and cannot directly represent contact force. All force-related analyses use calibrated values:

\begin{equation}
\Delta\mathbf{F}(t) = \mathbf{F}(t) - \mathbf{F}(0),
\label{eq:force_calibration}
\end{equation}

where \(t=0\) is the first static frame of each trajectory. The calibrated force reflects the actual force-change magnitude from the initial state to the current state and is more suitable for analyzing contact events and demonstration quality.

\section{Extended Limitations and Future Work}\label{app:future_work}

The main paper focuses on precision insertion because this setting requires tactile feedback to play multiple temporal roles simultaneously. However, PACE should be further evaluated on broader contact-rich manipulation tasks, including wiping, pushing and pulling, deformable-object manipulation, and bimanual assembly. These tasks may contain different contact-phase structures and different event time scales.

The current Event cue is built from short-window contact changes. This design is simple and effective for insertion, but it still uses fixed hyperparameters such as \(\alpha\), \(T_w\), and \(h_{\text{corr}}\). In future work, these time scales could be learned or conditioned on the estimated contact phase. For example, stable insertion may benefit from slower contact-state updates, while collision recovery may require a shorter event window and a sharper gate.

The current contact-consequence prediction head is used only during training. A natural extension is to use the predicted contact consequences at inference time. The policy could generate multiple candidate chunks, predict their likely contact changes, and select a safer action sequence before dispatch. This would turn Pred. from an auxiliary representation-learning signal into an anticipatory safety or planning signal.

PACE also assumes that the physical feedback signal is sufficiently informative. Cross-signal experiments show that GF225 tactile force, end-effector force/torque, and joint current can all support PACE, but weak contact, poor calibration, or sensor drift can still reduce reliability. More robust online calibration, uncertainty-aware event gating, or sensor-quality-aware fusion would make the method more robust in long-horizon deployment.

\end{document}